\documentclass[letterpaper, 10 pt, conference]{ieeeconf}  % Comment this line out
\IEEEoverridecommandlockouts                              % This command is only
\usepackage[utf8]{inputenc}
\usepackage{amsmath,amssymb,stmaryrd,mathtools}
\usepackage{amsfonts}
\usepackage[T1]{fontenc}
\usepackage[noend]{algpseudocode}
\usepackage{acronym}
\usepackage{verbatim}
\usepackage{booktabs}
\usepackage{siunitx}
\usepackage{graphics}
\usepackage{graphicx,caption}
\usepackage[keeplastbox]{flushend}
\usepackage{suffix}
\usepackage{xstring}
\usepackage{xparse}
\usepackage{expl3}
\usepackage{mathrsfs}
\usepackage{tabularx}
\usepackage{makecell}
\usepackage{array}
\usepackage{hyperref}

\usepackage{cleveref}
\usepackage[ruled,linesnumbered]{algorithm2e}
\usepackage{multirow}
\usepackage{diagbox}
\usepackage{rotating}
\usepackage{subcaption}
\usepackage{wrapfig}

\usepackage{tikz}
\usetikzlibrary{arrows,backgrounds,calc}
\usepackage{relsize}
\usepackage{float}
\usepackage{kantlipsum} 
\usepackage{lipsum}
\usepackage{stfloats}
\usepackage{siunitx}

\usepackage[noadjust]{cite}
\usepackage{todonotes}
\usepackage{soul}
\definecolor{smoothgreen}{rgb}{0.7,1,0.7}
\sethlcolor{smoothgreen}

\RequirePackage{luatex85}
\usepackage{pgfplots}
\pgfplotsset{compat=newest}
\pgfplotsset{every axis legend/.append style={%
		cells={anchor=west}}
}
\usepgfplotslibrary{polar}
\usetikzlibrary{arrows}
\tikzset{>=stealth'}

\definecolor{C1}{rgb}{0.0, 0.447, 0.741}
\definecolor{C1_light}{rgb}{0.0, 0.6032388663967612, 1.0}
\definecolor{C2}{rgb}{0.85, 0.325, 0.098}
\definecolor{C3}{rgb}{0.929, 0.694, 0.125}
\definecolor{C4}{rgb}{0.494, 0.184, 0.556}
\definecolor{C5}{rgb}{0.466, 0.674, 0.188}
\definecolor{C6}{rgb}{0.301, 0.745, 0.933}
\definecolor{C7}{rgb}{0.635, 0.078, 0.184}

\usepgfplotslibrary{groupplots}

\usetikzlibrary{shapes.geometric, arrows}

\tikzstyle{startstop} = [rectangle, rounded corners, minimum width=2cm, minimum height=1cm,text centered, draw=black, fill=none]
\tikzstyle{arrow} = [thick,->,>=stealth]

\title{
Learning to Navigate Intersections\\with Unsupervised Driver Trait Inference
}

\author{Shuijing Liu, Peixin Chang, Haonan Chen, \\ Neeloy Chakraborty, and Katherine Driggs-Campbell\thanks{S. Liu, P. Chang, H. Chen, N. Chakraborty and K. Driggs-Campbell are with the Department of  Electrical and Computer Engineering at the University of Illinois at Urbana-Champaign. emails: \{sliu105,pchang17,haonan2,neeloyc2,krdc\}@illinois.edu}
\thanks{This material is based upon work supported by the National Science Foundation under Grant No. 2143435.}}

\begin{document}
\maketitle
\thispagestyle{empty}
\pagestyle{empty}

%%%%%%%%%%%%%%%%%%%%%%%%%%%%%%%%%%%%%%%%%%%%%%%%%%%%%%%%%%%%%%%%%%%%%%%%%%%%%%%%
\begin{abstract}
Navigation through uncontrolled intersections is one of the key challenges for autonomous vehicles.
Identifying the subtle differences in hidden traits of other drivers can bring significant benefits when navigating in such environments.
We propose an unsupervised method for inferring driver traits such as driving styles from observed vehicle trajectories. 
We use a variational autoencoder with recurrent neural networks to learn a latent representation of traits without any ground truth trait labels.
Then, we use this trait representation to learn a policy for an autonomous vehicle to navigate through a T-intersection with deep reinforcement learning. 
Our pipeline enables the autonomous vehicle to adjust its actions when dealing with drivers of different traits to ensure safety and efficiency. Our method demonstrates promising performance and outperforms state-of-the-art baselines in the T-intersection scenario. For code implementation and videos, please visit {\color{cyan}{\url{https://github.com/Shuijing725/VAE_trait_inference}}}.

\end{abstract}

%%%%%%%%%%%%%%%%%%%%%%%%%%%%%%%%%%%%%%%%%%%%%%%%%%%%%%%%%%%%%%%%%%%%%%%%%%%%%%%%
\section{Introduction}
\label{sec:intro}

% Adapting to different types of human drivers is an essential capability for autonomous vehicles to successfully navigate through uncontrolled intersections~\cite{interactiondataset}. 
To successfully navigate through uncontrolled intersections, autonomous vehicles must carefully reason about how to interact with different types of human drivers~\cite{interactiondataset}. 
% The agent must balance safety and efficiency by being cautious when drivers are aggressive or irrational and bold when drivers passive or cooperative.
It is important for the vehicle to infer the traits of human drivers, such as the propensity for aggression or cooperation, and adjust its strategies accordingly~\cite{sunberg2017value,ma2020reinforcement}.
Inspired by recent advancements of unsupervised learning and deep reinforcement learning, we propose a novel pipeline to learn a representation of traits of other drivers, which is used for autonomous navigation in uncontrolled intersections.

% 1. need to infer driver internal states
% 2. fewer rules & sometimes rules are broken -> more complex & general decision making
Trait inference is challenging yet essential for the navigation of autonomous vehicles for two reasons. First, the environment is not fully observable to the ego vehicle, since each traffic participant runs its own policy individually and has its own internal states. The ego vehicle needs to interpret the hidden states such as driving styles and intents of other agents to understand future behaviors that may influence planning~\cite{sunberg2017value,ma2020reinforcement}. Second, uncontrolled intersections are less structured since traffic lights and stop signs are not present to coordinate agent behaviors. The traffic participants implicitly interact and negotiate with each other, making the environment complex and potentially dangerous~\cite{national2019traffic,qian2019toward,song2016intention,liu2020decentralized}. By inferring the traits of other drivers, the ego vehicle can be cautious when other drivers are aggressive or irrational and bold when they are passive or cooperative, improving both the safety and efficiency of interactive navigation.

% prev work: Morton et al, Ma et al and their problems

% To estimate the traits of drivers, Morton \textit{et al} uses an encoder to learn a latent representation of traits while simultaneously learns a policy that imitates the trajectories based on on the latent encodings and the driver states~\cite{morton2017simultaneous}. 
To address the above problems, 
Morton \textit{et al} learns a latent representation of driver traits, which is fed into a feedforward policy to produce multimodal behaviors~\cite{morton2017simultaneous}. However, the feedforward policy only considers current states and actions which are not sufficient to fully express long-term properties of drivers such as traits. As a result, this representation fails to distinguish between different traits.
Ma \textit{et al} classifies driver traits with supervised learning to aid navigation in intersections~\cite{ma2020reinforcement} but has the following two problems. First, the trait labels are expensive to obtain and usually do not exist in most real driving datasets~\cite{highDdataset, interactiondataset}. Second, the navigation policy is trained with ground truth trait labels instead of predicted traits. 
% The navigation policy takes ground truth trait labels as inputs in training. Thus, 
When the trait classifier and the policy are combined in testing, intermediate and cascading errors cause severe performance degradation.

\begin{figure}
    \centering
    \vspace{7pt}
    \includegraphics[scale=0.11]{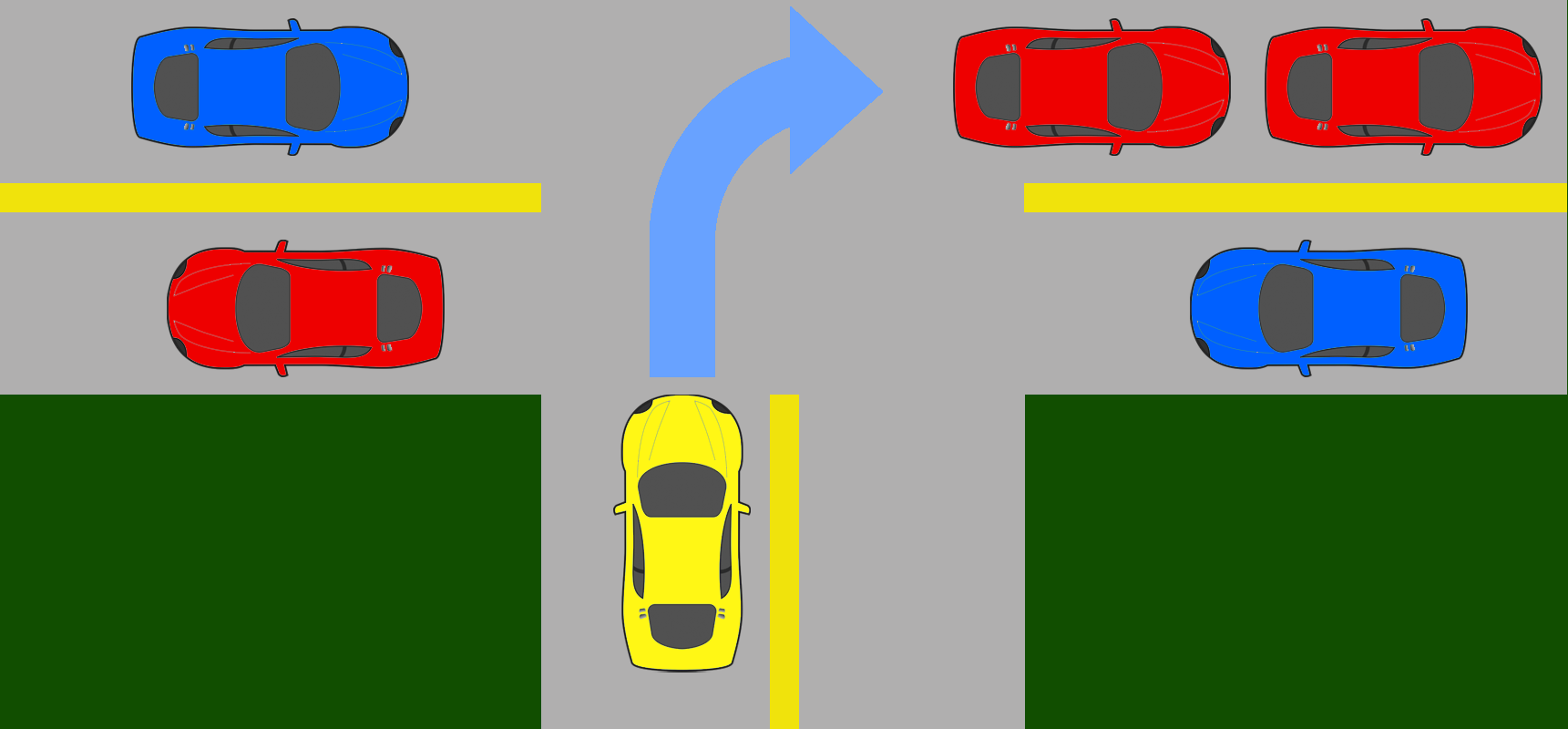}
    \caption{\textbf{The T-intersection scenario in left-handed traffic.} The goal of the ego car (yellow) is to take a right turn and merge into the upper lane without colliding with other cars. The conservative car (blue) yields to the ego car while the aggressive cars (red) do not yield.}
    \label{fig:opening}
    \vspace{-20pt}
\end{figure}

In this paper, we study the same uncontrolled T-intersection navigation problem as in \cite{ma2020reinforcement}, which is shown in Fig.~\ref{fig:opening}. 
Before entering the T-intersection, the ego vehicle needs to infer the latent driving styles of other drivers to determine whether they are willing to yield to the ego vehicle. Based on the inferred driving style, the ego vehicle must intercept the drivers who will yield to achieve the goal.

We seek to create a pipeline that first learns a representation of driver traits in an unsupervised way, and then uses the trait representation to improve navigation in the T-intersection.
% trait inference: method, network, and training
In the first stage, we encode the trajectories of other drivers to a latent representation using a variational autoencoder (VAE) with recurrent neural networks (RNN). Since trajectory sequences reveal richer information about driver traits than single states, the VAE+RNN network effectively learns to distinguish between different traits without any explicit trait labels.
% RL navigation
In the second stage, we use the latent representations of driver traits as inputs to the ego vehicle's navigation policy, which is trained with model-free reinforcement learning (RL).
With the inferred traits, the ego vehicle is able to adjust its decisions when dealing with different drivers, which leads to improved performance.
% advantages & how it addresses the problems of prev works
Since the RL policy is trained with the learned representations instead of ground truth labels, our pipeline is much less sensitive to cascading errors from the trait inference network. 

We present the following contributions: (1) We propose a novel unsupervised approach to learn a representation of driver traits with a VAE+RNN network; (2) We use the learned representation to improve the navigation of an autonomous vehicle through an uncontrolled T-intersection; (3) Our trait representation and navigation policy exhibit promising results and outperform previous works.

This paper is organized as follows: We review related works in Section~\ref{sec:related}. We formalize the problem and propose our method in Section~\ref{sec:methods}. Experiments and results are discussed in Section~\ref{sec:sim_exp}. We conclude the paper in Section~\ref{sec:conclusion}.

\section{Related Works}
\label{sec:related}

\subsection{Driver internal state estimation}
\label{sec:internal_state}
Driver internal state estimation can be divided into \textit{intent estimation} and \textit{trait estimation}~\cite{brown2020taxonomy}.
Intent estimation often uses methods such as probabilistic graph model and unparameterized belief tracker to predict the future maneuvers of other drivers~\cite{song2016intention,dong2017intention,bai2015intention}, which can then inform downstream planning for the ego driver~\cite{driggs2017integrating,song2016intention,bai2015intention}.
Trait estimation infers the properties of drivers such as driving styles, preferences, fatigue, and level of distraction~\cite{brown2020taxonomy}. Some works distinguish between distracted and attentive drivers for behavior prediction and cooperative planning~\cite{driggs2015improved,sadigh2018planning}. Driving style recognition has been addressed with both unsupervised and supervised learning methods, which we will discuss in detail below~\cite{martinez2017driving,dong2017autoencoder,morton2017simultaneous,ma2020reinforcement}. 

Morton \textit{et al} propose a method that first encodes driving trajectories with different driving styles to a latent space. Then, the latent encodings and the current driver states are fed into a feedforward policy that produces multimodal actions~\cite{morton2017simultaneous}.
The encoder and the policy are optimized jointly.
However, since the feedforward policy only considers the relations between current states and actions, the joint optimization encourages the encoder to encode short-term information of the trajectories such as accelerations while ignoring the persistent properties such as traits.
In contrast, since our method uses an RNN decoder to reconstruct the trajectories, the encoder must encode the trait information.
Ma \textit{et al} propose a graph neural network that classifies the driving styles with supervised learning, which requires large amounts of labeled data and is difficult to scale in reality~\cite{ma2020reinforcement}. In addition, since the definitions of trait properties vary by individuals and cultures, manually labeled trait information will likely be noisy and inconsistent. In contrast, our method is unsupervised and does not need any trait labels.

% VAE or self-supervised learning?
\subsection{Representation learning for sequential data}
% Social NCE & random swap of videos: negative sampling is task specific & in general tricky to design
Contrastive learning is widely used to learn representations from sequential data such as videos and pedestrian trajectories~\cite{wang2015unsupervised,misra2016shuffle,liu2020snce}. However, the performance of contrastive learning is sensitive to the quality of negative samples, and finding efficient negative sampling strategies remains an open challenge~\cite{grill2020bootstrap}.

Another category of representation learning methods is VAE and its variants~\cite{kingma2013auto,sohn2015learning}. Bowman \textit{et al} introduce an RNN-based VAE to model the latent properties of sentences~\cite{bowman2015generating}, which inspires us to learn the traits of drivers from their trajectories. Conditional VAEs (CVAE) are widely used in pedestrian and vehicle trajectory predictions since discrete latent states can represent different behavior modes such as braking and turning~\cite{salzmann2020trajectron,ivanovic2020multimodal,feng2019vehicle,schmerling2018multimodal}.
While these behavior modes change frequently, the driver traits that we aim to learn are persistent with each driver~\cite{brown2020taxonomy}. Also, the goal of these CVAEs is to generate multimodal trajectory predictions while, to the best of our knowledge, our work is the first to infer driver traits with VAE for autonomous driving.
% VAE: NLP paper & prediction papers (Trajectron ++, Tsigshua)
\subsection{Autonomous driving in uncontrolled intersections}
\label{sec:rl4driving}
Autonomous navigation through uncontrolled intersections is well studied and has had many successful demonstrations~\cite{isele2018navigating,song2016intention, cosgun2017towards,ma2020reinforcement}. 
% heuristic approach: TTC
Some heuristic methods use a time-to-collision (TTC) threshold to decide when to cross~\cite{minderhoud2001extended,cosgun2017towards}. However, the TTC models assume constant velocity for the surrounding vehicles, which ignores the interactions and internal states of drivers. Also, the TTC models can be overly cautious and cause unnecessary delays~\cite{isele2018navigating}.  
% imitation learning & pomdp methods
Another line of works formulates the problem as a partially observable Markov decision process (POMDP), which accounts for the uncertainties and partial observability in the intersection scenario but is computationally expensive to solve~\cite{song2016intention,bouton2017belief,brechtel2014probabilistic}. 
% RL based approach
RL-based methods use neural networks as function approximators to learn driving policies.
Isele \textit{et al} learns to navigate in occluded intersections using deep Q-learning ~\cite{isele2018navigating}. Ma \textit{et al} focuses on navigation in a T-Intersection where the drivers exhibit different traits~\cite{ma2020reinforcement}. However, the navigation policy is trained with ground truth traits and only uses the inferred traits in testing. Thus, the performance of the RL policy is sensitive to the intermediate errors from the trait inference module. In contrast, our method trains the RL policy directly with inferred trait representations, which eliminates the problems caused by the intermediate errors.

\section{Methodology}
In this section, we first present our unsupervised approach to represent driver traits from unlabeled driving trajectories with a VAE+RNN network. Then, we discuss how we use the inferred traits to improve the navigation policy.
\label{sec:methods}

% network architecture
\begin{figure*}[ht]
    \centering
    \includegraphics[width=\linewidth]{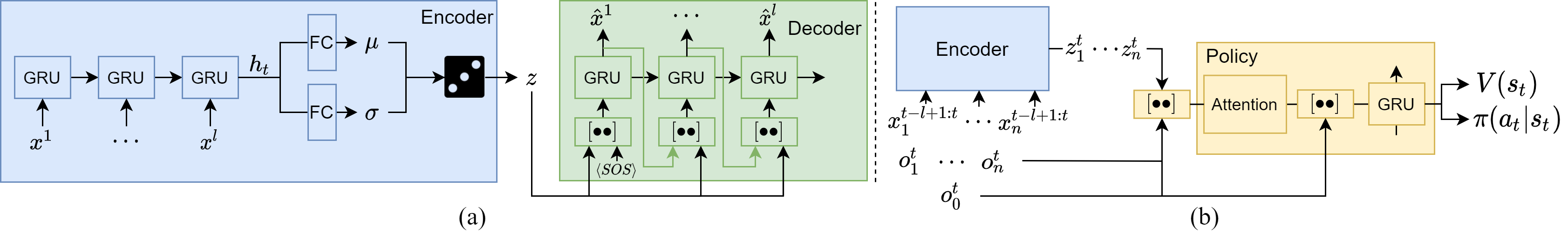}
    \caption{\textbf{The network architectures.} 
    (a) The VAE+RNN network for trait representation learning. The fully connected layers before and after GRUs are eliminated for clarity. The dice represents the reparameterization trick. The start-of-sequence state is denoted by $\langle SOS\rangle$. (b) The navigation policy network. The weights of the encoder (blue) is fixed and only the yellow part is trained with RL. We use $[\bullet \bullet]$ to denote concatenation. For illustration purposes, we assume that the inferred trait $z_1, ..., z_n$ is updated at time $t$.}
    \label{fig:network}
    \vspace{-15pt}
\end{figure*}

\subsection{Preliminaries}
Consider a T-intersection environment with an ego vehicle and $n$ surrounding vehicles. The number of surrounding vehicles $n$ may change at any timestep $t$. 
Suppose that all vehicles move in a 2D Euclidean space. Let ${o}_0^t$ be the state of the ego vehicle and ${o}_i^t$ be the observable state of the $i$-th surrounding vehicle at time $t$, where $i\in \{1, ..., n\}$. 
The state of the ego vehicle ${o}_0^t$ consists of the position $(p_x, p_y)$ and the velocity $(v_x, v_y)$ of the vehicle. The observable state of each surrounding vehicle ${o}_i^t$ consists of its position $(p_x^i, p_y^i)$. In contrast to the previous work~\cite{ma2020reinforcement}, ${o}_i^t$ does not include other vehicles' velocities because they are hard to accurately measure without special facilities and algorithms in the real world~\cite{lee2021study,han2017adaptive}. In addition, each surrounding vehicle has a latent state $z_i$ that indicates the aggressiveness (i.e., the trait) of the $i$-th driver. 
% Conservative cars on average have larger desired front gaps and larger desired speeds than aggressive cars. While conservative drivers would yield to the ego vehicle if the ego vehicle intercepts in the front, aggressive drivers would not. 
We assume that the driving style of each driver $z_i$ does not change throughout an episode.
The positions of surrounding vehicles ${o}_1^t, ..., {o}_n^t$ are observable to the ego vehicle, while the latent states $z_1, ..., z_n$ are not.

\subsection{Trait representation learning}
% goal
% In the first stage of our pipeline, we use a VAE+RNN network to learn a representation of driving styles from a set of unlabeled driver trajectories. 

\subsubsection{Data collection}
For each vehicle in the T-Intersection, the vehicle in front of it has the most direct influence on its behaviors. For this reason, in the trait representation learning stage, we define the observable state of each vehicle to be $x = (\Delta p_x, \Delta p_{x, f})$, where $\Delta p_x$ is the horizontal offset from the vehicle's starting position in the trajectory, and $\Delta p_{x, f}$ is the horizontal displacement of the vehicle from its front vehicle. Then, the trajectory of each driver is a sequence of states $\mathbf{x} = [x^1, ..., x^l]$, where $l$ is the length of the trajectory. 

We only keep the longitudinal state information because all vehicles move in horizontal lanes and the lateral states are not insightful in this setting, except indicating which lane the vehicle is in. We rotate the trajectories so that all trajectories in the dataset are aligned in the same direction. Thus, the lane and the directional information is indistinguishable within the trajectory data, allowing the network to focus on learning the trait difference instead of other differences between vehicles.

To collect the trajectory data, we run a simulation of the T-intersection scenario without the presence of the ego car and record the trajectories of all surrounding vehicles, which are controlled by the Intelligent Driver Model (IDM)~\cite{kesting2010enhanced}. 
Learning trait representations from this dataset allows the ego car to infer the traits from the trajectories of other drivers before deciding to intercept or wait. 
The dataset is denoted as $\{\mathbf{x}_j\}_{j=1}^N$, where $N$ is the total number of trajectories. 

\subsubsection{Network architecture}
We use the collected dataset to train the VAE+RNN network to learn a representation of traits, as shown in Fig.~\ref{fig:network}a. 
% network architecture 
% high level:
% encoder: p(z|traj), decoder: q(traj|z).
% The VAE network consists of an encoder $q_{\mathbf{\theta}}(z | \mathbf{x})$, which compresses a trajectory $\mathbf{x}$ to a distribution of latent trait vectors $z$, and a decoder, which reconstructs the trajectory from a conditional distribution $p_{\mathbf{\phi}}(\mathbf{x} | z)$. The objective of the VAE network is a lower bound on the marginal log likelihood of $\mathbf{x}$:
% \begin{equation}\label{eq:vae}
% \begin{aligned}
%     & \log p_{\mathbf{\theta}}(\mathbf{x}) \geq \mathcal{L}(\mathbf{\theta}, \mathbf{\phi}; \mathbf{x}) =\\
%      &-D_{KL}\left( q_{\mathbf{\theta}}(z | \mathbf{x}) || p_{\mathbf{\phi}}(\mathbf{x} | z)\right) + \mathbb{E}_{q_{\mathbf{\theta}}(z | \mathbf{x})}\left[p_{\mathbf{\phi}}(\mathbf{x} | z)\right]
% \end{aligned}
% \end{equation}
The VAE network consists of an encoder, which compresses a trajectory $\mathbf{x}$ to a distribution of a latent trait vector $z$, and a decoder, which reconstructs the trajectory from the latent vector $z$. 
Both the encoder and the decoder are gated recurrent unit (GRU) networks since GRUs are more computationally efficient than long short-term memory networks (LSTM). 

Given a trajectory $\mathbf{x} = [x^1, ..., x^l]$, the encoder GRU first applies a non-linear embedding layer $f_{\textrm{encoder}}$ to each state $x^t$ and then feeds the embedded features to the GRU cell:
\begin{equation}
	h^t_{e}=\mathrm{GRU}\left(h^{t-1}_{e}, f_{\textrm{encoder}}(x^t)\right)
\end{equation}
where $h^t_{e}$ is the hidden state of the encoder GRU at time $t\in\{1,..., l\}$. After the entire trajectory is fed through the encoder GRU, we take the last hidden state $h^l_{e}$ as the encoded latent feature of the trajectory $\mathbf{x}$ and apply fully connected layers $f_{\mu}$ and $f_{\sigma}$ to get the Gaussian parameters of the latent driving style $z\in \mathbb{R}^2$ in a two-dimensional latent space:
\begin{equation}
    \mu = f_{\mu}(h^t_{e}), \quad \sigma_i = f_{\sigma}(h^t_{e}).
\end{equation}
% Finally, we use the reparameterization trick to sample $z$ from $\mathcal{N}(\mu, \sigma)$ for efficient learning:
% \begin{equation}
%     z = \mu + \epsilon\sigma, \epsilon\sim \mathcal{N}(0, I)
% \end{equation}
Finally, we use the reparameterization trick to sample $z$ from $\mathcal{N}(\mu, \sigma)$ for efficient learning: $z = \mu + \epsilon\sigma, \epsilon\sim \mathcal{N}(0, I)$.

% decoder: concat z and x^t, eq for each step, how we handle the first step with a special SOT state, and how we end up with the reconstructed traj
In the decoding stage, since the driving style of each driver does not change over time, we treat the latent state $z$ as part of the vehicle state instead of the initial hidden state of decoder GRU. 
To this end, at each timestep $t$, we concatenate the reconstructed state $\hat{x}^{t-1}$ from the last timestep and the latent state $z$ from the encoder. Then, we embed the joint states using $f_{\textrm{decoder}}$, feed the embeddings into the next decoder GRU cell, and apply another embedding $g_{\textrm{decoder}}$ to the next hidden state $h^t_{d}$, which outputs the next reconstructed state $\hat{x}^{t}$:
\begin{equation}
\label{eq:decode}
\begin{aligned}
	h^t_{d} &=\mathrm{GRU}\left(h^{t-1}_{d}, f_{\textrm{decoder}}([\hat{x}^{t-1}, z])\right) \\
	\hat{x}^{t} &= g_{\textrm{decoder}}(h^t_{d}).
	\end{aligned}
\end{equation}
In the first timestep, we use a special start-of-sequence (SOS) state, which is similar to the start-of-sequence symbol in natural language processing to reconstruct $\hat{x}^1$~\cite{grefenstette2015learning}. The process in Eq.~\ref{eq:decode} repeats until we reconstruct the whole trajectory $\hat{\mathbf{x}}=[\hat{x}^1, ..., \hat{x}^l]$.

% training
The objective for training our VAE+RNN network is
\begin{equation}\label{eq:vae_rnn}
    \mathcal{L} = \beta D_{KL}\left(\mathcal{N}(\mu, \sigma) || \mathcal{N}(0, I)\right) + ||\mathbf{x} - \hat{\mathbf{x}}||_2
\end{equation}
where $D_{KL}$ is the Kullback–Leibler (KL) divergence. The first term regularizes the distribution of the latent trait $z$ to be close to a prior with a standard normal distribution. The second term is the reconstruction loss and measures the $L2$ error of the reconstructed trajectories from the original trajectories. The two terms are summed with a weight $\beta$.

By optimizing Eq.~\ref{eq:vae_rnn}, our network learns latent encodings that represent the trait of each trajectory without any ground truth trait labels. Note that we also make no assumptions on the number of trait classes or the semantic meanings of the trait classes. Thus, our network has the potential to generalize to real trajectory datasets with more complex traits.

\subsection{Navigation policy learning}
\label{sec:nav_policy}
% POMDP formulation, state & action representation, (briefly) transition & reward & value
We model the T-intersection scenario as a POMDP, defined by the tuple $ \langle \mathcal{S}, \mathcal{A}, \mathcal{T}, \mathcal{R}, \mathcal{O}, \mathcal{P}, \gamma \rangle$. Suppose that there are $n$ surrounding vehicles at timestep $t$. We use ${o}_t=[{o}_0^t, {o}_1^t, ..., {o}_n^t] \in \mathcal{O}$ to denote the observations of the ego vehicle, where $\mathcal{O}$ is the observation space. Let ${u}^t_i=[{o}^t_i, z_i]\in\mathbb{R}^4$ be the state of the $i$-th surrounding vehicle. Since the ego vehicle is influenced by all surrounding vehicles, we use $s_t=[{o}^t_0, {u}^t_1, ..., {u}^t_n] \in \mathcal{S}$ to denote the state of the POMDP, where $\mathcal{S}$ is the state space. And $\mathcal{P}:\mathcal{S}\rightarrow \mathcal{O}$ is the set of conditional observation probabilities.

At each timestep $t$, the ego vehicle chooses the desired speed for the low-level controller $a_t\in\mathcal{A}$ according to its policy $\pi(a_t|s_t)$, where $\mathcal{A}=\{0, 0.5, 3\}\mathrm{m/s}$ is the action space. In return, the agent receives a reward $r_t$ and transits to the next state $s_{t+1}$ according to an unknown state transition $\mathcal{T}(\cdot|s_t, a_t)$. Meanwhile, all other vehicles also take actions according to their policies and move to the next states with unknown state transition probabilities.
The episode continues until $t$ exceeds the maximum episode length $T$, the ego vehicle reaches its goal, or the ego vehicle collides with any other vehicle. The goal of the agent is to maximize the expected return, $R_t=\mathbb{E}[\sum^T_{i=t}\gamma^{i-t}r_{i}]$, where $\gamma$ is a discount factor. The value function $V^\pi (s)$ is defined as the expected return starting from $s$, and successively following policy $\pi$.

To handle the unknown transitions and environment complexity, we train our policy network illustrated in Fig.~\ref{fig:network}b using model-free deep RL. During RL training, we fix the trainable weights of the encoder. For every $l$ timesteps, we feed the trajectories of the surrounding vehicles over the past $l$ timesteps to the encoder and infer the driving style $z_i^t$ of each driver $i$. 
To improve efficiency, we only update the latent states of the drivers in the lanes that the ego car has not passed yet, as shown in Fig.~\ref{fig:qual_result}. Since the ego car is not allowed to move backward, the latent states of the drivers that the ego car has already passed or the drivers that have already yielded to the ego car have no effect on the ego car's decisions and thus are not updated anymore. 

% doge
% For the $i$-th driver, we concatenate together the inferred driving style $z_i$ , the observable states $\mathbf{o}_i$, and the state of the ego vehicle  $\mathbf{o}_0$. Let the $n$ concatenated vectors be denoted as $Q$, where $Q \in \mathbb{R}^{n \times 7}$. 
% v1
% We concatenate the state of each driver $\mathbf{u}^t_i$ with the state of the ego vehicle  $\mathbf{o}_0$ to obtain $Q = [[\mathbf{u}^t_1, \mathbf{o}_0];...; [\mathbf{u}^t_n, \mathbf{o}_0]]$, where $Q \in \mathbb{R}^{n \times 7}$. 
% We feed $Q$ into an attention module which assigns attention weights to each surrounding vehicle. Specifically, we embed $Q$ using a multi-layer perceptron (MLP) denoted as $f_{emb}$ and calculate the mean of the embeddings of each surrounding vehicle:
% \begin{equation}
% m=\frac{1}{n}\sum^n_{i=1} e_i, \quad E=f_{emb}(Q)
% \end{equation}
% where $m$ is the resulting mean and $e_i$ is the $i$-th row of $E$. The output of the attention module $c$  is calculated by
% v2
Our policy network is a GRU with an attention module. We concatenate the state of each driver ${u}^t_i$ with the state of the ego vehicle  ${o}_0$ to obtain ${q}_i^t = [{u}^t_i, {o}_0^t]$, where ${q}_i^t \in \mathbb{R}^{8}$ and $i\in\{1, ..., n\}$. 
We feed each concatenated state ${q}_i^t$ into an attention module which assigns attention weights to each surrounding vehicle. Specifically, we embed ${q}_i^t$ using a multi-layer perceptron (MLP) denoted as $f_{emb}$ and calculate the mean of the embeddings of each surrounding vehicle:
\begin{equation}
m^t=\frac{1}{n}\sum^n_{i=1} {e}_i^t, \quad {e}_i^t=f_{emb}({q}_i^t)
\end{equation}
where ${m}^t$ is the resulting mean. The weighted feature of each surrounding vehicle ${c}_i^t$ is calculated by
% doge
% \begin{equation}
% c=E^\top\alpha, \quad\alpha=f_{attn}([E,M]) 
% \end{equation}
% where $\alpha \in \mathbb{R}^{n \times 1}$ is the attention score for each surrounding vehicle, $f_{attn}$ is another MLP, and $M$ has $n$ rows, each of which is $m$. We then concatenate $c$ with $\mathbf{o}_0$ and feed the result to a GRU. 
\begin{equation}
{c}_i^t=\alpha_i^t \cdot {e}_i^t, \quad\alpha_i=f_{attn}([{e}_i^t,{m}^t]) 
\end{equation}
where $\alpha_i^t \in \mathbb{R}$ is the attention score for the $i$-th vehicle, and $f_{attn}$ is another MLP. We then concatenate the sum of $c_1^t, ..., c_n^t$ with the state of the ego vehicle ${o}_0^t$ and feed the result to a GRU:
\begin{equation}
    h^t_{\pi}=\mathrm{GRU}\left(h^{t-1}_{\pi}, \left[\sum^n_{i=1} c_i^t, o_0^t\right]\right)
\end{equation}
Finally, the hidden state of the GRU $h^t_{\pi}$ is fed to a fully connected layer to obtain the value $V(s_t)$ and the policy $\pi(a_t|s_t)$. 
We use Proximal Policy Optimization (PPO), a model-free policy gradient algorithm, for policy and value function learning~\cite{schulman2017proximal}.

\begin{figure*}[ht]
    \centering
    \includegraphics[scale=0.13]{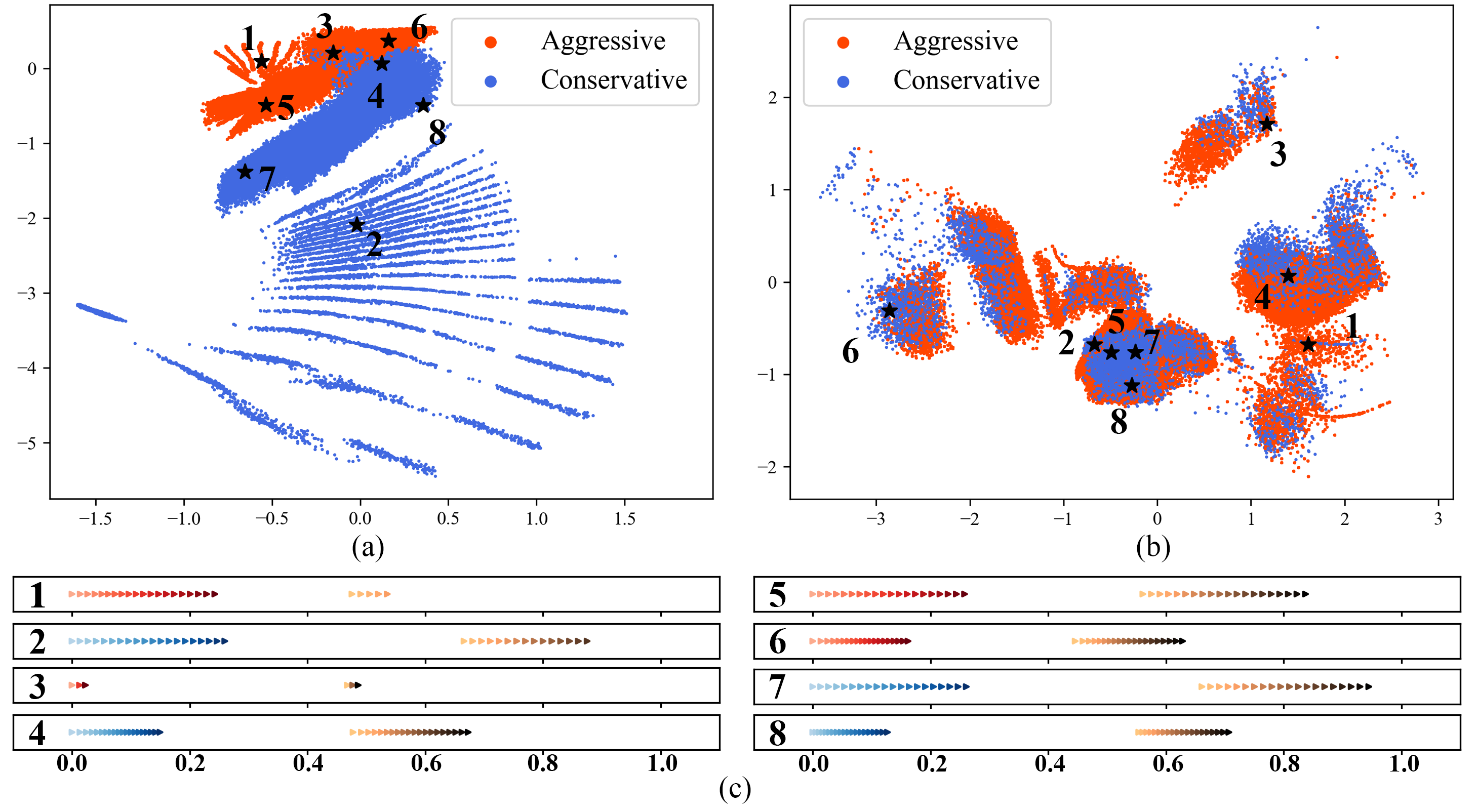}
    \vspace{-5pt}
    \caption{\textbf{Visualizations of the latent representations of two driver traits.} 
    (a) Our method. (b) The method by Morton \textit{et al.} (c) The original processed trajectories corresponding to the labeled latent vectors in (a) and (b). The $x$-axis is the horizontal displacement in meters. Each triangle marker indicates the position of a car at each timestep and the markers become darker over time. Denser triangles indicate smaller velocities. The blue and red trajectories indicate conservative and aggressive cars respectively. The brown trajectories indicate the front cars. If the front car goes out of the boundary before the trajectory ends, the brown trajectories will be shorter than the red or blue trajectories, such as \#1 and \#2.}
    \label{fig:vae_rep}
    \vspace{-15pt}
\end{figure*}

\section{Experiments and results}
\label{sec:sim_exp}

In this section, we first describe the simulation environment for training. We then present our experiments and results for trait representation and navigation policy. 
\subsection{Simulation environment}
Fig.~\ref{fig:qual_result} shows our simulation environment adapted from~\cite{ma2020reinforcement}. At the beginning of an episode, the surrounding vehicles are randomly placed in a two-way street with opposite lanes and we assume that they never take turns or change lanes. 
The number of surrounding vehicles varies as vehicles enter into or exit from the T-intersection. We assume that all cars can always be detected and tracked. 
The surrounding vehicles are controlled by IDM~\cite{kesting2010enhanced}. 
Conservative drivers vary their front gaps from the preceding vehicles between $0.5m$ and $0.7m$ and have the desired speed of $2.4m/s$. Aggressive drivers vary their front gaps between $0.3m$ and $0.5m$ and have the desired speed of $3m/s$. If the ego car moves in front of other cars, conservative drivers will yield to the ego car, while aggressive drivers will ignore and collide with the ego car. 

The ego car starts at the bottom of the T-intersection. The goal of the ego car is to take a right turn to merge into the upper lane without colliding with other cars. The ego car is controlled by a longitudinal PD controller whose desired speed is set by the RL policy network. The right-turn path of the ego car is fixed to follow the traffic rule. The ego car also has a safety checker that clips the magnitude of its acceleration if it gets dangerously close to other cars.

Let $S_{goal}$ be the set of goal states, where the ego vehicle successfully makes a full right-turn, and $S_{fail}$ be the set of failure states, where the ego vehicle collides with other vehicles. Let  $r_{speed}(s)=0.05\times \lVert v_{ego} \rVert_2$ be a small reward on the speed of the ego vehicle and $r_{step}=-0.0013$ be a constant penalty that encourages the ego vehicle to reach the goal as soon as possible. The reward function is defined as
\begin{equation}
\label{eq:reward}
 \begin{split}
\begin{gathered}
    r(s, a)  = 
        \begin{cases}
            2.5, & \text{if } s\in S_{goal}\\
            -2, & \text{if } s\in S_{fail}\\
            r_{speed}(s)+r_{step}, & \text{otherwise}.
        \end{cases}
\end{gathered}
\end{split}
\end{equation}

\subsection{Trait inference}

\subsubsection{Baseline}
We compare the latent representation of our method with the method proposed by Morton \textit{et al}~\cite{morton2017simultaneous}. 
\begin{table}[h]
  \begin{center}
    \caption{Testing results of simple supervised classifiers with learned latent representations}
    \label{tab:classification}
    \begin{tabular}{ll} 
    \toprule
    \textbf{Method} & Accuracy \\
      \midrule
      \textbf{Ours} & \textbf{98.08\%}\\
      \textbf{Morton \textit{et al.}} & 60.22\%\\
      \bottomrule
    \end{tabular}
  \end{center}
  \vspace{-22pt}
\end{table}
The encoder of the baseline is the same as our encoder except that the baseline takes the longitudinal acceleration at each timestep as an extra input. The policy network of the baseline is a 4-layer multilayer perceptron (MLP) network that imitates the IDM policy from the trajectory dataset.

\subsubsection{Training and evaluation} 
Our dataset contains approximately $696, 000$ trajectories from both classes, and the train/test split ratio is $2$:$1$. We train both methods for $1000$ epochs with a decaying learning rate $5\times10^{-4}$. The weight of the KL divergence loss $\beta$  is $5\times10^{-8}$ for both methods.

To better understand the learned latent representation and how it provides trait information to the navigation policy, we fix the trainable weights of the encoders, train linear support vector classifiers using the inferred latent states as inputs, and record the testing accuracy of the classifiers. 
To visualize the learned representations, we feed a set of testing trajectories into the encoder and visualize their latent representations. 
\subsubsection{Results and interpretability}
% As shown in Table~\ref{tab:classification}, the supervised classifier with our latent representation achieves much higher classification accuracy than that of Morton \textit{et al}. This gap indicates that our VAE+RNN network produces a better representation of driver traits, which has a higher potential to benefit downstream tasks such as navigation.   

% our method: two traits are mostly separated 
In Table~\ref{tab:classification}, the supervised classifier with our latent representation achieves much higher classification accuracy than that of Morton \textit{et al}.
Together with Fig.~\ref{fig:vae_rep}a, we show that our representation successfully separates the vehicle trajectories with different traits in most cases. 
For example, in Fig.~\ref{fig:vae_rep}c, \#1, \#5, and \#6 are in the aggressive cluster, while \#2, \#7, and \#8 are in the conservative cluster.
% the overlaps contain ambiguous trajectories such as very short traj and the traj with front gap (and velocity maybe) in between 
The overlapped region encodes the trajectories with ambiguous traits, such as very short trajectories (\#3) and the trajectories with ambiguous front gaps. For example, the average front gap of \#4 is between the aggressive \#6 and the conservative \#8.
% besides trait, our latent representation also maps traj with different velocities and those with disappeared front car to a meaningful position
Besides the binary traits, our latent representation also captures other meaningful information. First, the trajectories of the cars whose front cars go out of the border, such as \#1 and \#2, are mapped into the fan-shaped peripherals of the two clusters respectively. Second, in the central regions of the two clusters, the trajectories with larger average speeds are mapped in the lower left part (\#5 and \#7), while those with smaller average speeds are in the upper right part (\#6 and \#8).
% TODO: why our method works
% Thus, by encoding and reconstructing the processed trajectories, our VAE+RNN network is able to effectively represent the differences among drivers including but not limited to traits.

% In contrast, two traits in Morton collapsed together, although it can distinguish long/short traj, traj with front car disappeared 
From Table~\ref{tab:classification} and Fig.~\ref{fig:vae_rep}b, the baseline suffers from severe model collapse and the representation fails to separate the two traits. The reason is that the MLP policy only considers current state-action pairs, which encourages the encoder to only encode features within short time windows such as accelerations while ignoring the properties of the trajectories such as traits.
Despite the model collapse, the baseline still learns some meaningful representations. For example, the trajectories with uniform speeds together such as \#5 and \#7 and forms separate clusters for the decelerating trajectories such as \#6. Therefore, we conclude that compared with single states, trajectories exhibit richer information about traits and are better suited for trait representation learning.
\subsection{Navigation with inferred traits}

\subsubsection{Baselines and ablations}
We use the following two baselines: (1) The pipeline proposed by Ma \textit{et al}, which trains a supervised trait predictor and an RL policy with binary ground truth trait labels separately and combines them at test time~\cite{ma2020reinforcement}; (2) We use the latent representation by Morton \textit{et al} to train a policy network using the same method described in Sec.~\ref{sec:nav_policy}. In addition, we use an RL policy trained with ground truth labels as an oracle, and another RL policy trained without any trait information as a na\"ive model. To examine the effectiveness of the attention mechanism, we train an ablated model of our method without the attention module. For a fair comparison, the architectures of all policy networks are kept the same.
\subsubsection{Training}
\begin{figure*}[ht]
    \centering
    \includegraphics[scale=0.33]{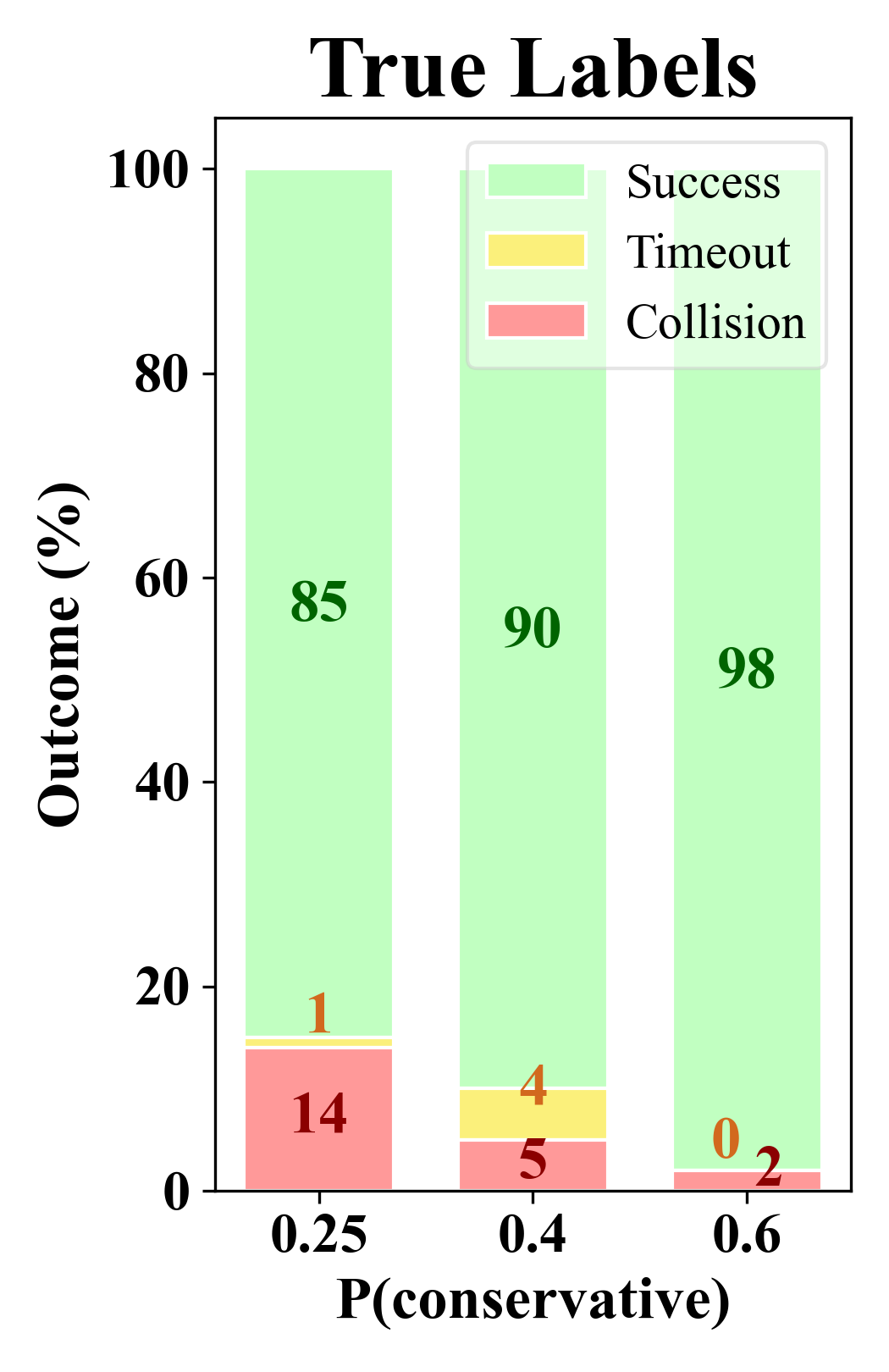}
    \includegraphics[scale=0.33,trim={1cm 0 0 0},clip]{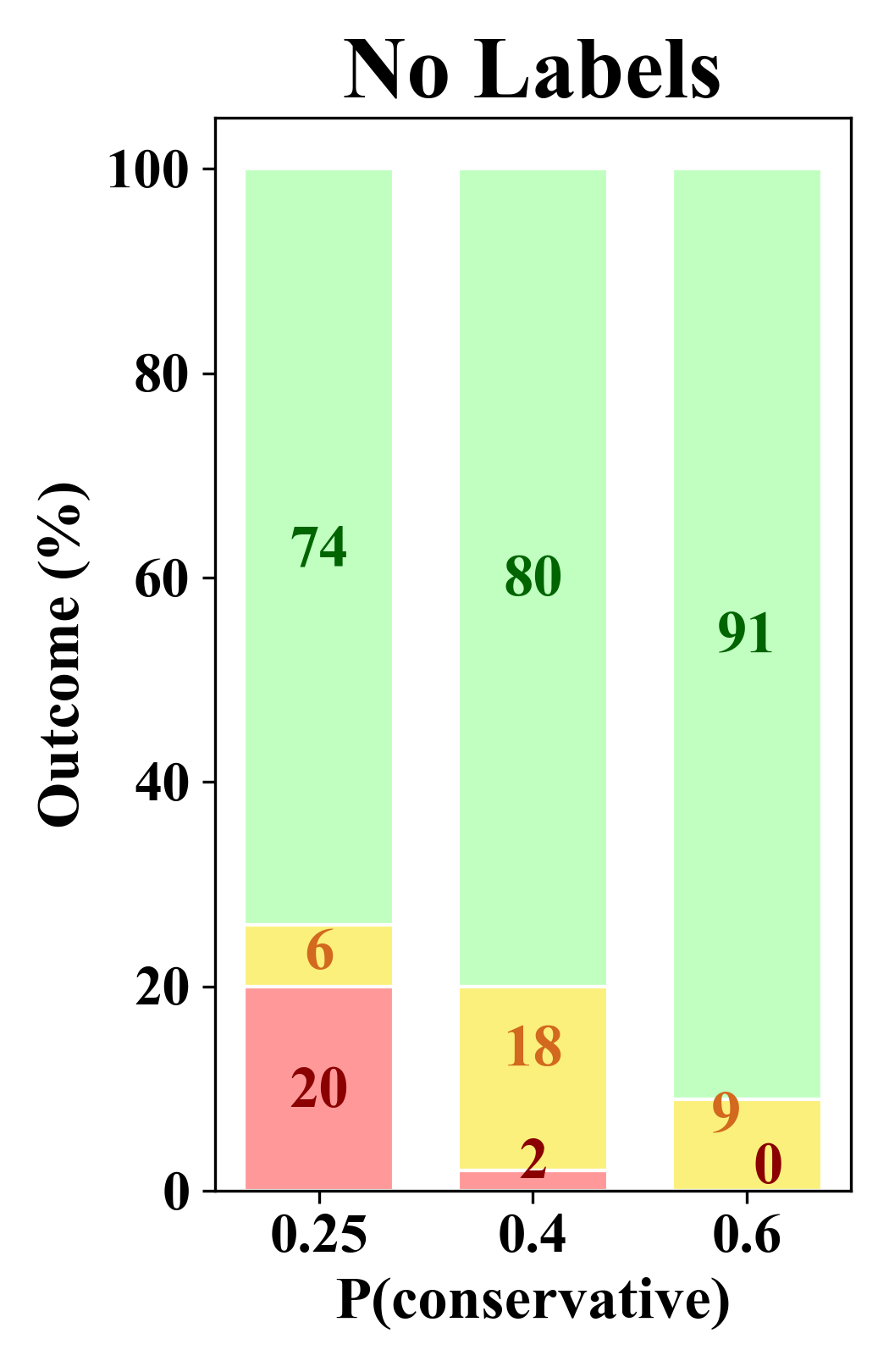}
    \includegraphics[scale=0.33,trim={1cm 0 0 0},clip]{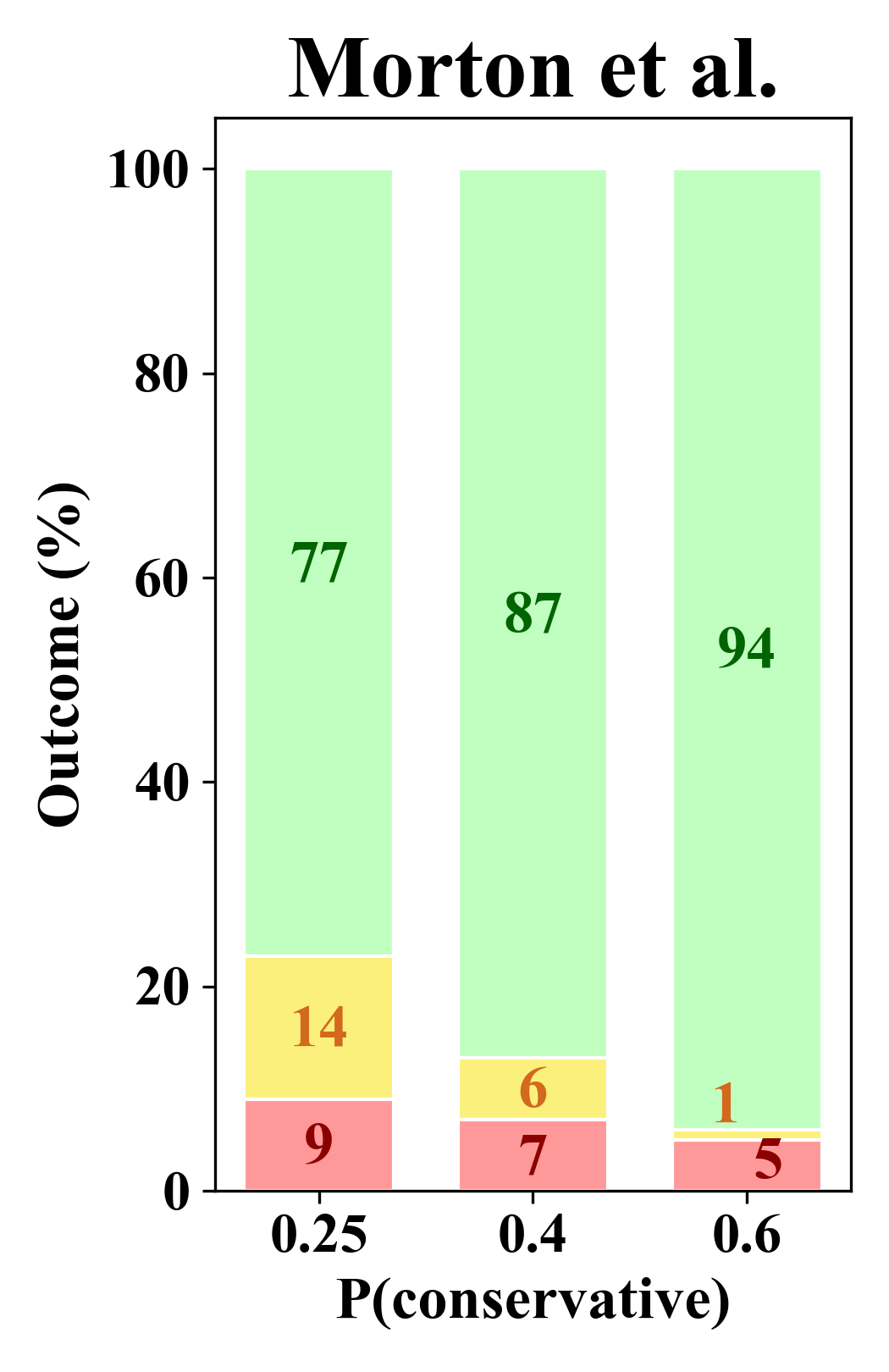}
    \includegraphics[scale=0.33,trim={1cm 0 0 0},clip]{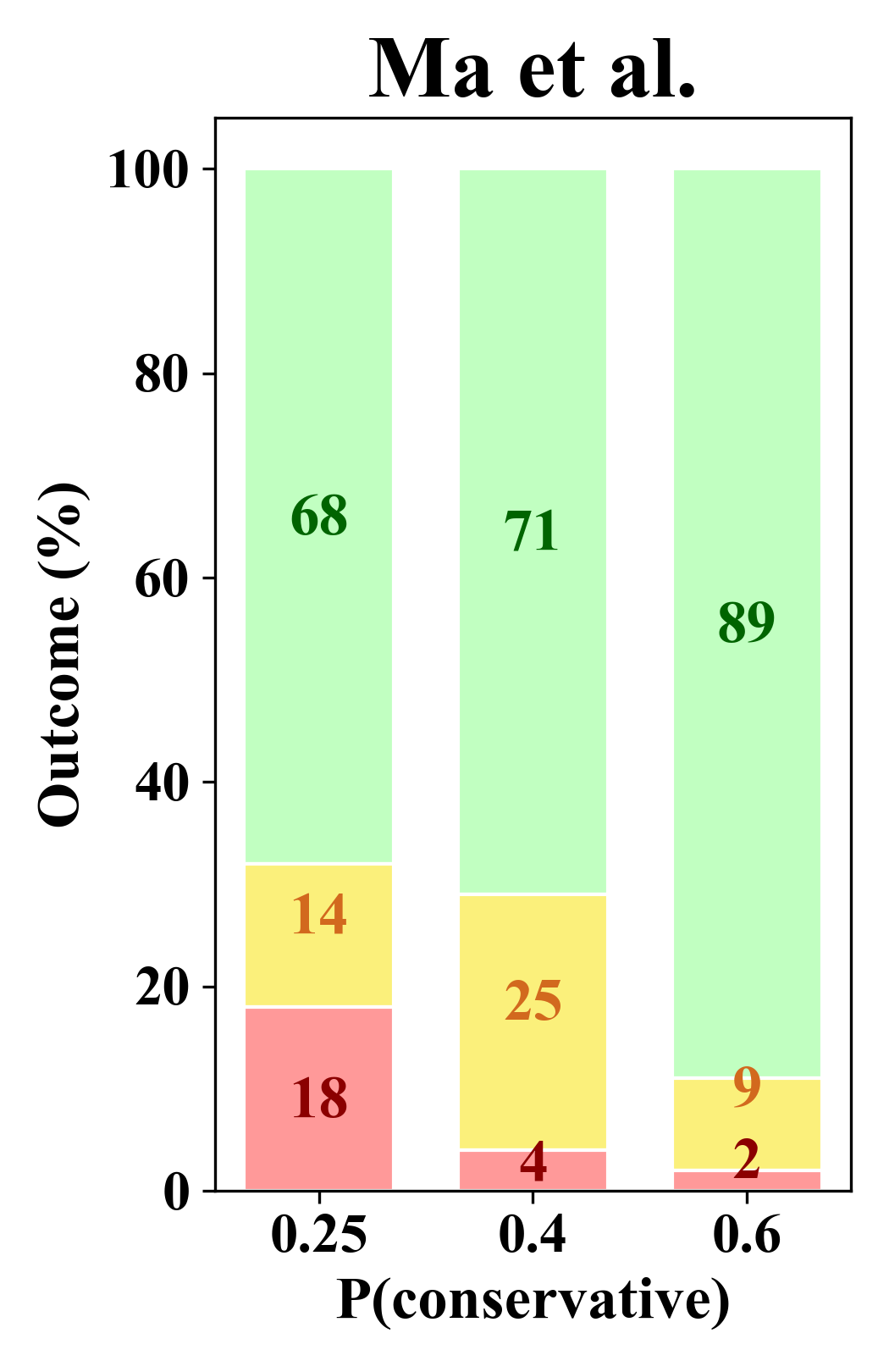}
    \includegraphics[scale=0.33,trim={1cm 0 0 0},clip]{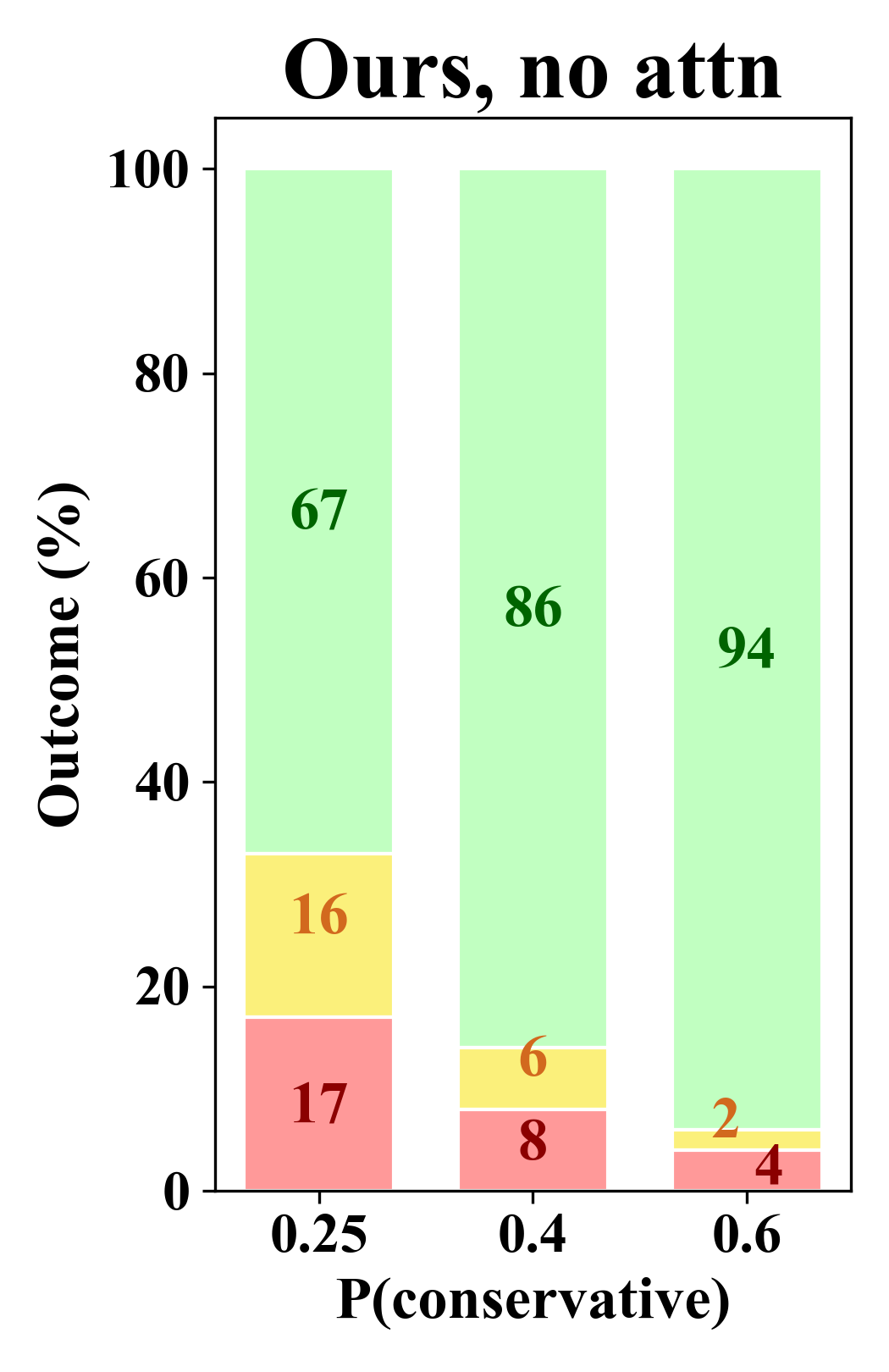}
    \includegraphics[scale=0.33,trim={1cm 0 0 0},clip]{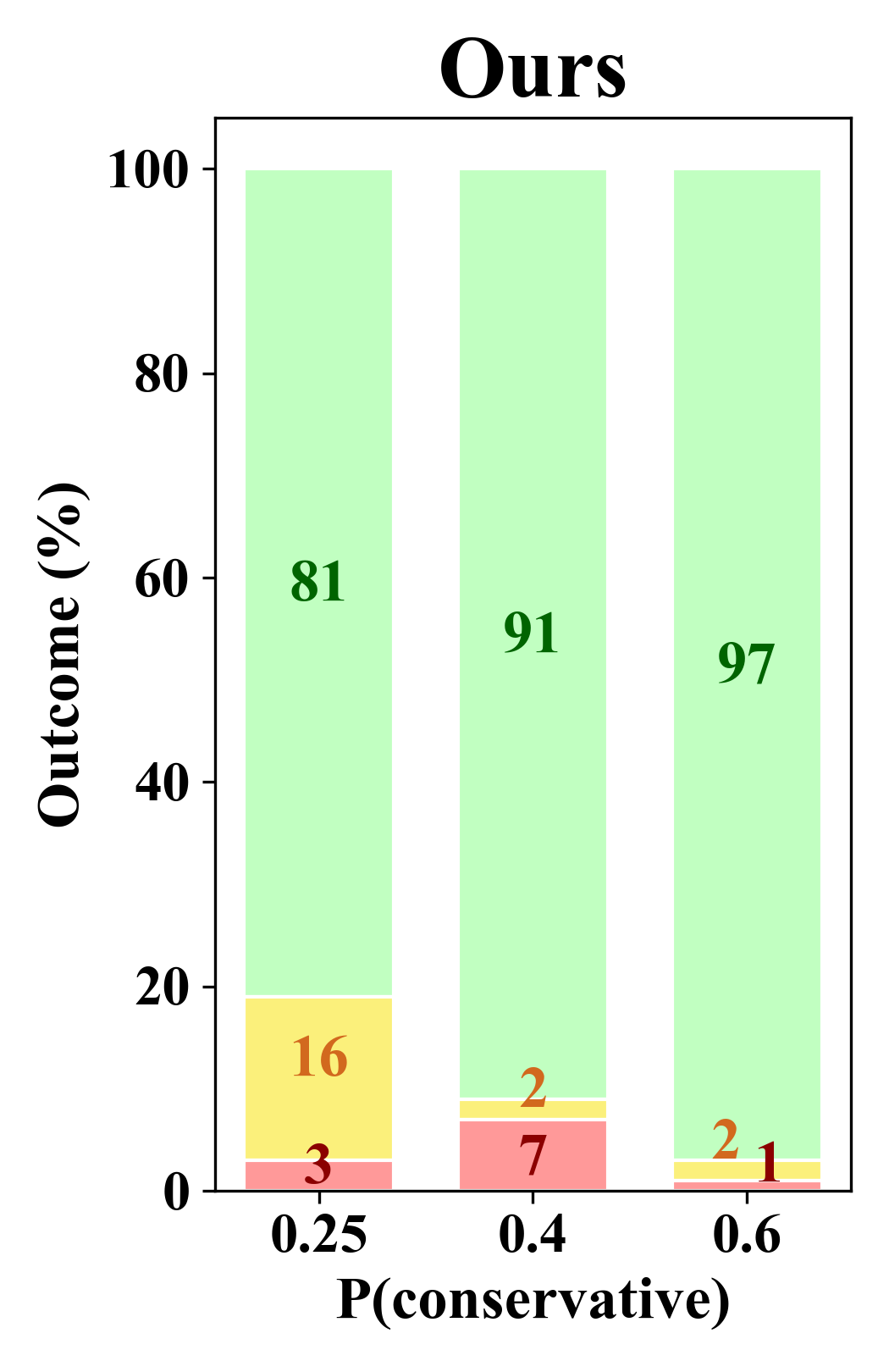}
    
    \vspace{-5pt}
    
    \caption{\textbf{Success, timeout, and collision rates w.r.t. different driver trait distributions.} $P(\textrm{conservative})$ is the probability for each surrounding driver to be conservative. The task difficulty increases as $P(\textrm{conservative})$ decreases. The numbers on the bars indicate the percentages of the corresponding bars.}
    \label{fig:quan_result}
    \vspace{-5pt}
\end{figure*}

We run three experiments with different proportions of two types of drivers, as shown in Fig.~\ref{fig:quan_result}. We train the policy networks for all methods and ablations for $1\times 10^7$ timesteps with a decaying learning rate $ 1\times 10^{-4}$. To accelerate and stabilize training, we run twelve instances of the simulation environment in parallel for collecting the ego car's experiences. At each policy update, 30 steps of 6 episodes are used.
For Ma \textit{et al}, we pretrain a trait classification network with a $96\%$ testing accuracy and use the classifier to infer traits for the RL policy. 

\subsubsection{Evaluation}
We test all models with $500$ random unseen test cases. We measure the percentage of success, collision, and timeout episodes as the evaluation criteria. 

\begin{figure*}[ht]
    \centering
     \includegraphics[scale=0.22]{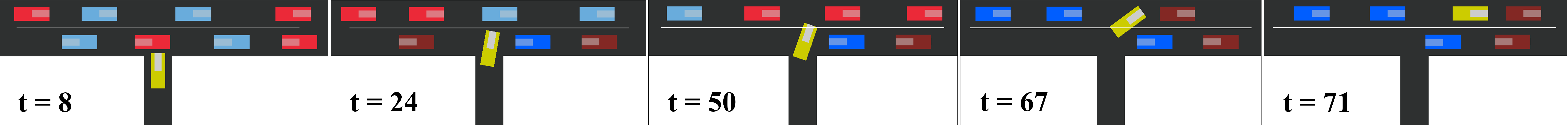}
    \caption{\textbf{Qualitative result of our method.} The ego car is in yellow, the conservative cars are in blue, and the aggressive cars are in red. As mentioned in Sec.~\ref{sec:nav_policy}, the latent traits of the light blue and light red cars are updated periodically, while those of the dark blue and dark red cars are not updated and stay the same as before. 
    }
    \label{fig:qual_result}
    \vspace{-20pt}
\end{figure*}

\subsubsection{Results}
% Compare with true & no labels
% true labels
As shown in Fig.~\ref{fig:quan_result}, the success rates of our method are closest to the oracle who has access to true trait labels in all experiments, with an average difference of $2\%$. 
We believe the main reason is that our trait representation effectively captures the trait differences of the surrounding drivers and aids the decision-making in RL. The performance gap between the oracle and our method is caused by the drivers with ambiguous traits, as well as the fact that the learned representation is noisier than the true labels.
% no labels: can implicitly infer but with our representation, the RL policy can learn whether to intercept other cars or not based on latent encodings instead of trajectories -> efficient 
Although the model with no labels can implicitly infer traits to some extent in RL training, our policy is able to utilize the existing trait representation and focuses more on the decision-making of the ego vehicle, which leads to better navigation performance.
Fig.~\ref{fig:qual_result} shows a successful episode of our method, where the ego car waits until a conservative car appears, cuts in the front of the conservative cars when passing both lanes, and completes the right turn.

% compare with baselines
% Morton: latent representation does not distinguish between traits -> the useful information for RL is limited, still need to implicitly infer 
Compared with our model, the model by Morton \textit{et al} has a lower success rate especially in more challenging experiment settings when $P(\textrm{conservative})$ is smaller. The reason is that the latent representation does not distinguish between different traits and only provides very limited useful information to RL. This implies that the policy still needs to implicitly infer the traits while being distracted by the latent representation, which negatively affects the performance.  

% Ma et al: The two modules has good performance when testing separately, but when combined together, intermediate and cascading errors since RL is trained with true labels, if latent inference module makes a small mistake, RL is sensitive to this mistake and fails easily
% As a result, although the trait inference module has a high testing accuracy, the policy fails easily whenever the inferred traits are wrong.
For Ma \textit{et al}, the trait classifier and the RL policy both have good performance when tested separately. However, when the two modules are combined together, intermediate and cascading errors significantly lower the success rates. The performance drop is due to the distribution shift between the true traits and inferred traits. Since the policy is trained with true traits, it fails easily whenever the trait classifier makes a small mistake. Moreover, Ma \textit{et al} requires trait labels, but our trait representation is learned without any labels and still outperforms Ma \textit{et al} in navigation.

\subsection{Attention vs. no attention}
The second from the rightmost graph in Fig.~\ref{fig:quan_result} shows that the removal of the attention module results in $3\%\sim14\%$ lower success rates. Since the attention module assigns different weights to the cars with different traits and in different positions, the policy is able to focus on the cars which have a bigger influence on the ego car, which validates the necessity of the attention mechanism.

% reason: attention module assigns different attention scores to cars with different traits, and the cars in different relative positions with the ego car -> the policy can focus on the cars which have a bigger influence on the ego car at each timestep -> validates necessity of attention layers

% \input{Sections/06-RealExp.tex}
% % \input{Sections/06-Results.tex}

\section{Conclusions and future work}
\label{sec:conclusion}
We propose a novel pipeline that encodes the trajectories of drivers to a latent trait representation with a VAE+RNN network. Then, we use the trait representation to improve the navigation of an autonomous vehicle through an uncontrolled T-intersection. 
The trait representation is learned without any explicit supervision.
Our method outperforms baselines in the navigation task and the trait representation shows interpretability. 
Possible directions to explore in future work include (1) validating our method with real driving trajectory data, (2) generalizing our model to more sophisticated driver internal states and more navigation tasks, and (3) incorporating occlusions and limited sensor range of the ego vehicle to close the gap between the simulation and the real world.

\section*{Acknowledgements}
% We thank Zhengguan Dai for setting up the baseline code and Zhe Huang for feedback on paper drafts.
We thank Xiaobai Ma for thoughtful discussions and for providing the baseline code. We thank Zhe Huang and Tianchen Ji for feedback on paper drafts.
%%%%%%%%%%%%%%%%%%%%%%%%%%%%%%%%%%%%%%%%%%%%%%%%%%%%%%%%%%%%%%%%%%%%%%%%%%%%%%%%


%%%%%%%%%%%%%%%%%%%%%%%%%%%%%%%%%%%%%%%%%%%%%%%%%%%%%%%%%%%%%%%%%%%%%%%%%%%%%%%%

\newpage
\clearpage
\bibliographystyle{IEEEtran}
\bibliography{BibFile}
\clearpage
\end{document}